\documentclass[sigconf]{acmart}
\usepackage{booktabs} 

\usepackage[english]{babel}
\usepackage{moresize}
\usepackage{amsmath}
\usepackage{algorithmic}
\usepackage{balance}
\usepackage{comment}
\usepackage{paralist}
\usepackage{bm}
\usepackage{pgfplots}
\usetikzlibrary{pgfplots.dateplot}

\usepackage{flushend}
\usepackage[english]{babel}
\usepackage[latin1]{inputenc}
\usepackage{mathrsfs}
\usepackage{graphicx}

\usepackage{amssymb}
\usepackage{amsfonts}
\usepackage{url}
\usepackage{longtable}
\usepackage{rotating}
\usepackage{multirow}
\usepackage{mathrsfs}
\usepackage{subfigure}
\usepackage{enumitem}
\usepackage[linesnumbered,algoruled,boxed,lined]{algorithm2e}
\usepackage{adjustbox}
\usepackage{hyperref}
\usepackage{pgfplots}
\usetikzlibrary{pgfplots.dateplot}
\usepackage{filecontents}

\definecolor{tblue}{RGB}{31,119,180}
\definecolor{torange}{RGB}{255,127,14}
\definecolor{tgreen}{RGB}{44,160,44}
\definecolor{tred}{RGB}{214,39,40}
\definecolor{tpurple}{RGB}{148,103,189}

\newcommand{\hide}[1]{} 

\newcommand{\eg}{\textit{e}.\textit{g}.}

\def\model{UrbanGPT}

\setcopyright{none}

\begin{document}
\fancyhead{}

\title{UrbanGPT: Spatio-Temporal Large Language Models}

\author{Zhonghang Li$^{1,2}$, Lianghao Xia$^1$, Jiabin Tang$^1$, Yong Xu$^2$, \\Lei Shi$^3$, Long Xia$^3$,  Dawei Yin$^3$ and Chao Huang$^{1*}$}
\thanks{$*$ Chao Huang is the Corresponding Author.}
\affiliation{$^1$The University of Hong Kong, $^2$South China University of Technology, $^3$Baidu Inc.\\}
\affiliation{\textbf{Project Page}: \href{https://urban-gpt.github.io/}{https://urban-gpt.github.io/}, \textbf{Github}: \href{https://github.com/HKUDS/UrbanGPT}{https://github.com/HKUDS/UrbanGPT}}


\begin{abstract}
Spatio-temporal prediction aims to forecast and gain insights into the ever-changing dynamics of urban environments across both time and space. Its purpose is to anticipate future patterns, trends, and events in diverse facets of urban life, including transportation, population movement, and crime rates. Although numerous efforts have been dedicated to developing neural network techniques for accurate predictions on spatio-temporal data, it is important to note that many of these methods heavily depend on having sufficient labeled data to generate precise spatio-temporal representations. Unfortunately, the issue of data scarcity is pervasive in practical urban sensing scenarios. In certain cases, it becomes challenging to collect any labeled data from downstream scenarios, intensifying the problem further. Consequently, it becomes necessary to build a spatio-temporal model that can exhibit strong generalization capabilities across diverse spatio-temporal learning scenarios. 

Taking inspiration from the remarkable achievements of large language models (LLMs), our objective is to create a spatio-temporal LLM that can exhibit exceptional generalization capabilities across a wide range of downstream urban tasks. To achieve this objective, we present the \model, which seamlessly integrates a spatio-temporal dependency encoder with the instruction-tuning paradigm. This integration enables LLMs to comprehend the complex inter-dependencies across time and space, facilitating more comprehensive and accurate predictions under data scarcity. To validate the effectiveness of our approach, we conduct extensive experiments on various public datasets, covering different spatio-temporal prediction tasks. The results consistently demonstrate that our \model, with its carefully designed architecture, consistently outperforms state-of-the-art baselines. These findings highlight the potential of building large language models for spatio-temporal learning, particularly in zero-shot scenarios where labeled data is scarce. 
\end{abstract}




\maketitle

\section{Introduction}
\label{sec:intro}

Spatio-temporal prediction is driven by the motivation to accurately forecast and gain valuable insights into the dynamic nature of urban environments. By analyzing and understanding the ever-changing dynamics across both time and space, spatio-temporal prediction allows us to anticipate future patterns, trends, and events in various aspects of urban life. This has significant implications in the field of urban computing, where the ability to predict transportation patterns can optimize traffic flow, reduce congestion, and enhance overall urban mobility~\cite{liao2018deep,wang2023pattern}. Moreover, anticipating population movement can aid in effective urban planning and resource allocation~\cite{feng2018deepmove,luca2021survey}. Additionally, the ability to forecast crimes can greatly contribute to enhancing public safety~\cite{wang2016crime}. Spatio-temporal prediction plays a vital role in shaping smarter and more efficient cities, ultimately leading to an improved quality of urban life.

It is important to highlight the various types of neural network architectures commonly adopted in this domain of spatio-temporal prediction. These architectures are designed to capture and model the complex relationships between spatial and temporal dimensions in the data. One widely employed architecture is the Convolutional Neural Network (CNN)~\cite{zhang2017deep,yao2018deep,li2022spatial}, which is effective in extracting spatial features by applying convolutional filters across the input data. Another line of spatio-temporal neural networks is the Recurrent Neural Network (RNN) family~\cite{bai2020adaptive,wang2020traffic,yu2017deep}. Those spatio-temporal RNNs are well-suited for capturing temporal dependencies by maintaining a memory state that can retain information over time. Recently, there has been a surge in the use of Graph Neural Networks (GNNs) for spatio-temporal prediction~\cite{wu2020connecting,Zhao2020TGCN,ye2021coupled}. GNNs excel in modeling complex spatial relationships in data represented as graphs, where each node corresponds to a spatial location and edges capture the connections between them.

While current spatio-temporal neural network techniques have proven to be highly effective, it is crucial to acknowledge their strong dependence on having an abundance of labeled data in order to generate accurate predictions. However, the pervasive problem of data scarcity in practical urban sensing scenarios poses a significant challenge. For example, deploying sensors throughout the entire urban space to monitor citywide traffic volume or air quality is impractical due to the high cost involved~\cite{yi2016st,liang2019urbanfm}. Moreover, the challenge of limited labeled data availability extends to spatio-temporal forecasting across different cities, in which acquiring labeled data for each target city becomes a daunting task~\cite{yao2019learning,jin2022selective}. These issues emphasize the pressing need for novel solutions that address data scarcity and enhance the generalization capabilities of spatio-temporal models in various smart city applications.

Inspired by the remarkable progress of large language models (LLMs), our primary goal is to create a spatio-temporal LLM that possesses outstanding generalization capabilities across a wide array of urban tasks. Leveraging the reasoning abilities inherent in LLMs, we aim to expand their success into the domain of spatio-temporal analysis. Our objective is to develop a model that can effectively comprehend and forecast intricate spatial and temporal patterns, enabling it to excel in various urban scenarios.

While it is of utmost importance to develop a versatile spatio-temporal model capable of effectively handling diverse downstream tasks, aligning the spatio-temporal context with the knowledge space of large language models (LLMs) and enabling them to comprehend the complex dependencies across time and space present significant challenges. These hurdles call for meticulous model design to bridge the gap between the unique characteristics of spatio-temporal data and the knowledge encoded within LLMs. \\\vspace{-0.12in}

\noindent \textbf{Contributions}. In light of these challenges, we propose \model, a large language model specifically tailored for spatio-temporal prediction. At the core of \model\ lies a novel spatio-temporal instruction-tuning paradigm that seeks to align the intricate dependencies of time and space, with the knowledge space of LLMs. Within our \model\ framework, we start by incorporating a spatio-temporal dependency encoder, which utilizes a multi-level temporal convolutional network. This encoder enables the model to capture the intricate temporal dynamics present in the spatio-temporal data across various time resolutions. Then, our model involves aligning textual and spatio-temporal information to empower language models in effectively injecting spatio-temporal contextual signals. This is achieved through the utilization of a lightweight alignment module that projects the representations of spatio-temporal dependencies. The result is the generation of more expressive semantic representations by integrating valuable information from both textual and spatio-temporal domains.

\begin{figure}
\centering
\subfigure{
    \begin{minipage}[t]{1\linewidth}
        \centering
        \includegraphics[width=3.19in]{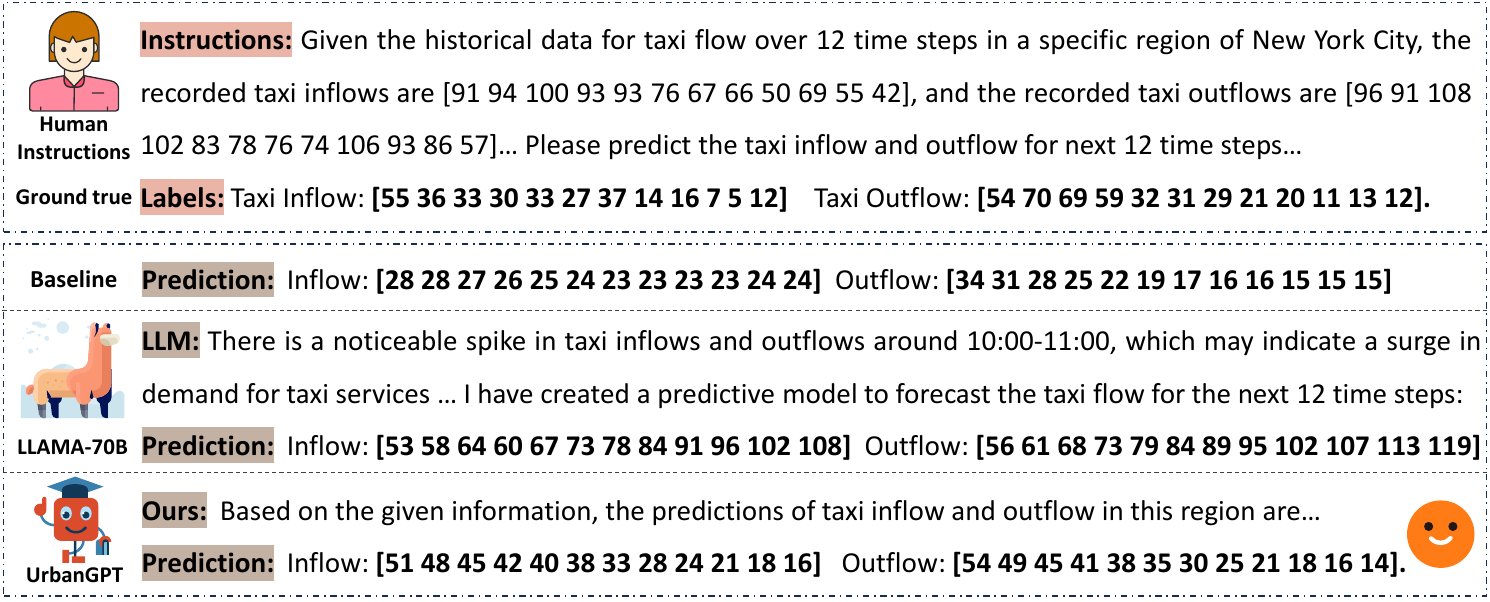}
    \end{minipage}%
}%
\\
\vspace{-0.1in}
\subfigure{
    \begin{minipage}[t]{0.48\linewidth}
        \centering
        \includegraphics[width=1.50in]{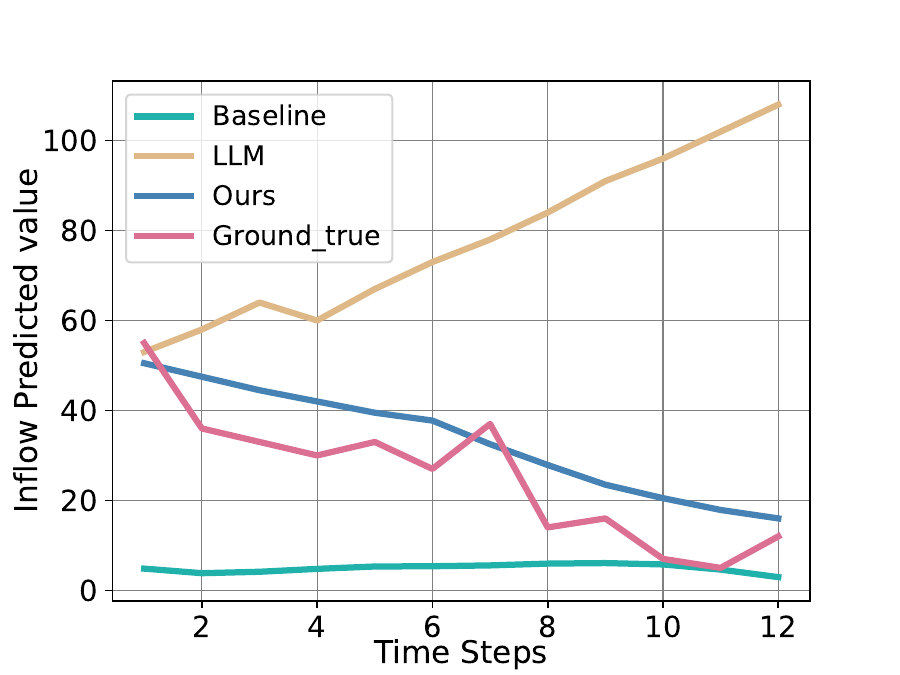}
    \end{minipage}%
    \begin{minipage}[t]{0.52\linewidth}
        \centering
        \includegraphics[width=1.50in]{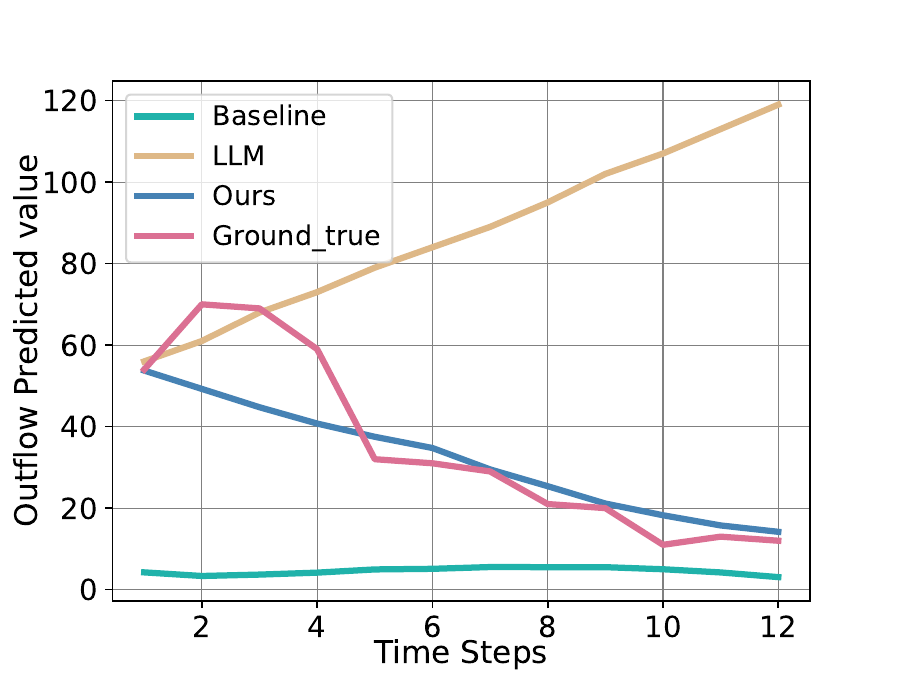}
    \end{minipage}%
}%
\centering
\vspace{-0.2in}
\caption{The superior predictive performance of the proposed \model\ compared to the large language model (LLaMA-70B) and the spatiotemporal graph neural network (STGCN) in a zero-shot traffic flow prediction scenario.}
\vspace{-0.25in}
\label{fig:intro}
\end{figure}

Through the incorporation of spatio-temporal information during the instruction-tuning process, the language model gains proficiency in understanding and processing the intricate relationships and patterns found in spatio-temporal data. By leveraging the insights obtained from the spatio-temporal domain, the language model becomes better equipped to capture the nuances and complexities of spatio-temporal phenomena. This, in turn, enables the model to make more reliable and insightful predictions across various urban scenarios, even when faced with limited data availability.

To showcase the superior predictive performance of our proposed model, we compare it with the large language model (LLaMA-70B) and the spatio-temporal graph neural network (STGCN) in a zero-shot traffic flow prediction scenario guided by textual instructions, as depicted in Figure~\ref{fig:intro}. The large language model, LLaMA, effectively infers traffic patterns from the input text. However, its limitations in handling numeric time-series data with complex spatial and temporal dependencies can sometimes lead to opposite traffic trend predictions. On the other hand, the pre-trained baseline model demonstrates a strong understanding of spatio-temporal dependencies. However, it may suffer from overfitting to the source dataset and underperform in zero-shot scenarios, indicating its limited generalization capabilities beyond existing spatio-temporal prediction models. In contrast, our proposed model achieves a harmonious integration of domain-specific spatio-temporal knowledge and language modeling capabilities. This enables us to make more accurate and reliable predictions under data scarcity.

In summary, our main contributions can be outlined as follows:
\begin{itemize}[leftmargin=*]

\item To the best of our knowledge, this is the first attempt to develop a spatio-temporal large language model capable of predicting diverse urban phenomena across different datasets, especially under conditions of limited data availability. 

\item We propose \model, a spatio-temporal prediction framework that empowers large language models (LLMs) to comprehend the intricate inter-dependencies across time and space. This is achieved through the seamless integration of a spatio-temporal dependency encoder with the instruction-tuning paradigm, effectively aligning the spatio-temporal context with LLMs.

\item Extensive experiments conducted on three benchmark datasets provide compelling evidence of our proposed \model's exceptional ability to generalize in zero-shot spatio-temporal learning scenarios. These findings highlight the model's robust generalization capacity, demonstrating its effectiveness in accurately predicting and understanding spatio-temporal patterns, even in scenarios where no prior training data is available.

\end{itemize}

\section{Preliminaries}
\label{sec:model}
\textbf{Spatio-Temporal Data}. 
Spatio-temporal data is commonly collected and can be represented as a three-dimensional tensor $\textbf{X}\in\mathbb{R}^{R\times T\times F}$. Each element $\textbf{X}_{r,t,f}$ in the tensor corresponds to the value of the $f$-th feature at the $t$-th time interval in the $r$-th region. To provide an example, let's consider predicting taxi traffic patterns in an urban area. In this scenario, the data can represent the inflow and outflow of taxis in a specific region (\eg, the $r$-th spatial area) during a given time period from $t$ to $t-1$ (\eg, a 30-minute interval). \\\vspace{-0.12in}

\noindent \textbf{Spatio-Temporal Forecasting.}
In spatio-temporal prediction tasks, a common scenario involves forecasting future trends using historical data. Specifically, the goal is to predict the data for the next $P$ time steps based on information from the preceding $H$ steps.
\begin{align}
    \label{eq:Preliminarie1}
    \textbf{X}_{t_{K+1}: t_{K+P}} = f(\textbf{X}_{t_{K-H+1}: t_K})
\end{align}
The function $f(\cdot)$ represents a spatio-temporal prediction model that has been trained effectively using historical data. Spatio-temporal prediction tasks can be divided into two main categories: regression prediction, which involves predicting continuous values like traffic flow or taxi demand~\cite{pan2019urban}, and classification prediction, where the goal is to classify events such as crime occurrence prediction~\cite{huang2018deepcrime}. To optimize the model $f(\cdot)$, different loss functions are utilized based on the specific characteristics of the spatio-temporal scenarios. \\\vspace{-0.12in}

\noindent \textbf{Spatio-Temporal Zero-Shot Learning}.
Despite the effectiveness of current spatio-temporal learning approaches, they often encounter difficulties in effectively generalizing across a wide range of downstream spatio-temporal learning scenarios. In this study, our focus is on addressing the challenge of spatio-temporal zero-shot scenarios, where we aim to learn from previously unseen data in downstream spatio-temporal prediction datasets or tasks. This can be formally defined as follows:
\begin{align}
    \label{eq:Preliminarie2}
    \tilde{\textbf{X}}_{t_{K+1}: t_{K+P}} = \hat{f}(\tilde{\textbf{X}}_{t_{K-H+1}: t_K})
\end{align}
In this particular scenario, the prediction function $\hat{f}(\cdot)$ is responsible for forecasting the spatio-temporal data $\tilde{\textbf{X}}$ from downstream tasks that have not been previously encountered. It should be noted that the model $\hat{f}(\cdot)$ is not trained specifically on the target data.

\section{Methodology}
\label{sec:solution}

\begin{figure*}
    \centering
    \includegraphics[width=2.1\columnwidth]{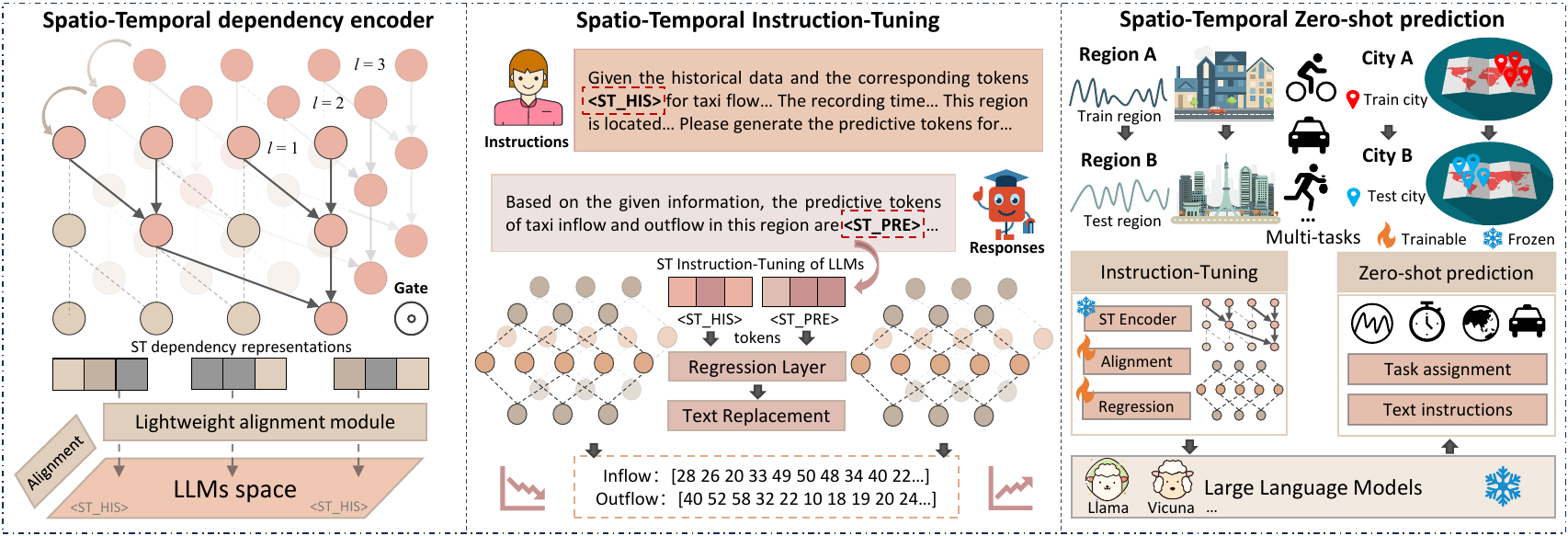}
    \vspace{-0.2in}
    \caption{The overall architecture of the proposed spatio-temporal language model \model.}
    \vspace{-0.1in}
    \label{fig:fra01}
\end{figure*}

\subsection{Spatio-Temporal Dependency Encoder}
Although large language models demonstrate exceptional proficiency in language processing, they face challenges in comprehending the time-evolving patterns inherent in spatio-temporal data. To overcome this limitation, we propose enhancing the capability of large language models to capture temporal dependencies within spatio-temporal contexts. This is accomplished by integrating a spatio-temporal encoder that incorporates a multi-level temporal convolutional network. By doing so, we enable the model to effectively capture the intricate temporal dependencies across various time resolutions, thereby improving its understanding of the complex temporal dynamics present in the spatio-temporal data. Specifically, our spatio-temporal encoder is composed of two key components: a gated dilated convolution layer and a multi-level correlation injection layer. Let's formalize this architecture as:
\begin{align}
    \label{eq:st_encoder1}
    \mathbf{\Psi}_{r}^{(l)} = (\bar{\textbf{W}}_k^{(l)} * \textbf{E}_{r}^{(l)} + \bar{\textbf{b}}_k^{(l)}) \cdot \delta(\bar{\textbf{W}}_g^{(l)} * \textbf{E}_{r}^{(l)} + \bar{\textbf{b}}_g^{(l)}) + \textbf{E}_{r}^{'(l)}
\end{align}
We begin with the initial spatio-temporal embedding, denoted as $\textbf{E}_{r}\in\mathbb{R}^{T\times d}$. This embedding is obtained by enhancing the original data $\textbf{X}$ through a linear layer. To address the issue of gradient vanishing, we utilize a slice of $\textbf{E}_{r}$, denoted as $\textbf{E}_{r}^{'}\in\mathbb{R}^{T^{'}\times d}$, which is determined by the size of the dilated convolutional kernel. This slice is employed for performing residual operations. To perform the residual operations, we use 1-D dilated convolution kernels $\bar{\textbf{W}}_k$ and $\bar{\textbf{W}}_g\in\mathbb{R}^{T_g\times d_{in}\times d_{out}}$, along with the corresponding bias terms $\bar{\textbf{b}}_k$ and $\bar{\textbf{b}}_g\in\mathbb{R}^{d_{out}}$. The sigmoid activation function $\delta$ is applied to control the degree of information preservation during the repeated convolution operation. After the gated temporal dilated convolutional layer encoding, we are able to effectively capture the temporal dependencies across multiple time steps, resulting in the generation of temporal representations. 

These representations contain different levels of temporal dependencies, reflecting various granularity-aware time-evolving patterns. To preserve these informative patterns, we introduce a multi-level correlation injection layer. This layer is designed to incorporate correlations between different levels and is formalized as:
\begin{align}
    \label{eq:st_encoder2}
    \textbf{S}_{r}^{(l)} &= (\textbf{W}_s^{(l)} * \mathbf{\Psi}_{r}^{(l)} + \textbf{b}_s^{(l)}) + \textbf{S}_{r}^{(l-1)}
\end{align}
We have the convolution kernel $\textbf{W}_s\in\mathbb{R}^{T_S\times d_{out}\times d_{out}^{'}}$ and the bias vector $\textbf{b}_s\in\mathbb{R}^{d_{out}^{'}}$. These are employed after $L$ layers of encoding. A simple non-linear layer 
is employed to merge the results from equations~\ref{eq:st_encoder1} and ~\ref{eq:st_encoder2}, and the final spatio-temporal dependency representations of are denoted as $\tilde{\mathbf{\Psi}}\in\mathbb{R}^{R\times F\times d}$. To address the diverse set of urban scenarios that may arise downstream, our proposed spatio-temporal encoder is designed to be independent of graph structures when modeling spatial correlations. This is particularly crucial because in zero-shot prediction contexts, the spatial relationships between entities may be unknown or difficult to ascertain. By not relying on explicit graph structures, our encoder can effectively handle a broad spectrum of urban scenarios, where spatial correlations and dependencies can vary or be challenging to define in advance. This flexibility enables our model to adapt and perform well, ensuring its applicability in a wide range of urban contexts.

\subsection{Spatio-Temporal Instruction-Tuning}
\label{sec:instruction-tune}

\subsubsection{\bf Spatio-Temporal-Text Alignment}
In order to enable language models to effectively comprehend spatio-temporal patterns, it is crucial to align textual and spatio-temporal information. This alignment allows for the fusion of different modalities, resulting in a more informative representation. By integrating contextual features from both textual and spatio-temporal domains, we can capture complementary information and extract higher-level semantic representations that are more expressive and meaningful. To achieve this objective, we utilize a lightweight alignment module to project the spatio-temporal dependencies representations $\tilde{\mathbf{\Psi}}$. This projection involves the use of parameters $\textbf{W}_p\in\mathbb{R}^{d\times d_L}$ and $\textbf{b}_p\in\mathbb{R}^{d_L}$, where $d_L$ represents the commonly used hidden dimension in language models (LLMs).

The resulting projection, denoted as $\textbf{H}\in\mathbb{R}^{R\times F \times d_L}$, are represented in the instructions using special tokens as: <ST\_start>, <ST\_HIS>, ..., <ST\_HIS>, <ST\_end>. Here, <ST\_start> and <ST\_end> serve as identifiers marking the beginning and end of the spatio-temporal token. These identifiers can be included in the large-scale language model by expanding its vocabulary. The placeholder <ST\_HIS> represents the spatio-temporal token and corresponds to the projection $\textbf{H}$ in the hidden layer. By employing this technique, the model gains the ability to discern spatio-temporal dependencies, thereby enhancing its proficiency in successfully performing spatio-temporal predictive tasks within urban scenes.

\subsubsection{\bf Spatio-Temporal Prompt Instructions}
\label{sec:st-context}
In scenarios involving spatio-temporal prediction, both temporal and spatial information contain valuable semantic details that contribute to the model's understanding of spatio-temporal patterns within specific contexts. For instance, traffic flow in the early morning differs significantly from rush hour, and there are variations in traffic patterns between commercial and residential areas. As a result, we recognize the potential of representing both temporal and spatial information as prompt instruction text. We leverage the text understanding capabilities of large language models to encode this information, enabling associative reasoning for downstream tasks.

\begin{figure}
    \centering
    \includegraphics[width=1\columnwidth]{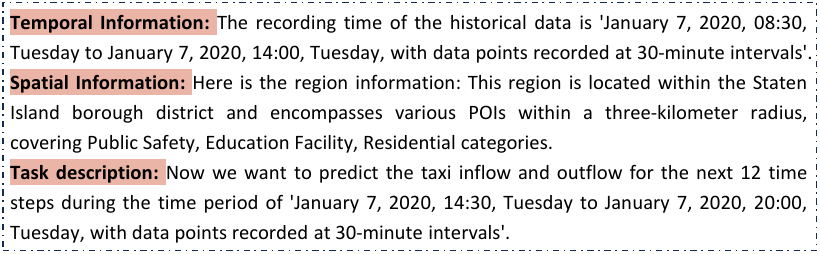}
    \vspace{-0.2in}
    \caption{Illustration of spatio-temporal prompt instructions encoding the time- and location-aware information.}
    \vspace{-0.1in}
    \label{fig:instruction}
\end{figure}

In our \model\ framework, we integrate multi-granularity time information and spatial details as instruction inputs for the large language model. Time information includes factors such as the day of the week and the hour of the day, while regional information encompasses the city, administrative areas, and nearby points of interest (POI) data, among others. By incorporating these diverse elements, \model\ is capable of identifying and assimilating spatio-temporal patterns across different regions and timeframes. This enables the model to encapsulate these insights within complex spatio-temporal contexts, thereby enhancing its ability for zero-shot reasoning. The design of the instructional text for spatio-temporal information is illustrated in Figure~\ref{fig:instruction}.

\subsubsection{\bf Spatio-Temporal Instruction-Tuning of LLMs}
\label{sec:Urban_prediction} 
When it comes to incorporating detailed spatio-temporal textual descriptions, the next stage is to fine-tune large language models (LLMs) using instructions to generate spatio-temporal forecasts in textual format. However, this approach poses two challenges. \emph{Firstly}, spatio-temporal forecasting typically relies on numerical data, which differs in structure and patterns from natural language that language models excel at processing, focusing on semantic and syntactic relationships. \emph{Secondly}, large language models are typically pre-trained using a multi-classification loss to predict vocabulary, resulting in a probability distribution of potential outcomes. This contrasts with the continuous value distribution required for regression tasks.

To address these challenges, \model\ adopts a different strategy by refraining from directly predicting future spatio-temporal values. Instead, it generates forecasting tokens that aid in the prediction process. These tokens are subsequently passed through a regression layer, which maps the hidden representations to generate more accurate predictive values. The formulation of the regression layer in our spatio-temporal instruction-tuning paradigm is:
\begin{align}
    \label{eq:regression_layer}
    \hat{\textbf{Y}}_{r,f} = \textbf{W}_3[\sigma(\textbf{W}_1 \textbf{H}_{r,f}), \sigma(\textbf{W}_2 \mathbf{\Gamma}_{r,f})]
\end{align} 
The prediction result, denoted as $\hat{\textbf{Y}}_{r,f}\in\mathbb{R}^{P}$, is obtained using the rectified linear unit activation function, represented by $\sigma$. The hidden representations of the forecasting tokens, denoted as $\mathbf{\Gamma}_{r,f}\in\mathbb{R}^{d_L}$, are introduced as a novel term in the vocabulary of large language models (LLMs). The regression layer is formulated using weight matrices $\textbf{W}_1\in\mathbb{R}^{d^{'}\times d_L}$, $\textbf{W}_2\in\mathbb{R}^{d^{'}\times d_L}$, and $\textbf{W}_3\in\mathbb{R}^{P\times 2d^{'}}$, where $[\cdot,\cdot]$ represents the concatenation operation. While the probability distribution of the forecasting tokens remains relatively stable, their hidden representations contain rich spatio-temporal contextual attributes that capture dynamic spatio-temporal interdependencies. This enables our model to provide precise predictions by leveraging this contextual information.

\subsection{\bf Model Optimization}
\label{sec:loss_function}
Building upon the baseline model~\cite{bai2020adaptive,Guo2019attention}, we adopt the absolute error loss as our regression loss function. This choice allows us to effectively handle predictions across a wide range of urban scenarios. Additionally, we introduce a classification loss as a joint loss to cater to diverse task requirements. To ensure optimal performance, our model optimizes different losses based on the specific task inputs. For instance, we utilize the regression loss for tasks such as traffic flow prediction, while employing the classification loss for tasks like crime prediction. This approach enables our model to effectively address the unique challenges posed by each task and deliver accurate predictions in various urban scenarios.
\begin{align}
    \label{eq:loss}
    \mathcal{L}_c &= -\frac{1}{N} \sum_{i=1}^{N} \left[\delta(y_i) \cdot \log(\hat{y}_i) + (1 - \delta(y_i)) \cdot \log(1 - \hat{y}_i) \right]\nonumber\\
    \mathcal{L}_r &= \frac{1}{N}\sum_{i=1}^N|y_i - \hat{y}_i|;~~~ \mathcal{L} = \mathcal{L}_{LLMs} + \mathcal{L}_r + \mathcal{L}_c
\end{align}
Here, $y_i$ represents a sample from $\hat{\textbf{Y}}$, and $N$ represents the total number of samples, which is calculated as the product of $R$, $T$, and $F$. We use various loss functions in our model, including $\mathcal{L}_c$ for binary cross-entropy loss, $\mathcal{L}_r$ for regression loss, and the cross-entropy loss adopted in our spatio-temporal language models. To capture probability distributions from the prediction, we employ the sigmoid function denoted by $\delta$. Each of these loss functions plays a specific role in our model, allowing us to effectively handle classification, regression, and language modeling tasks as needed.

\section{Evaluation}
\label{sec:eval}

In this section, we aim to assess the capabilities of our proposed model across various settings by addressing five key questions:
\begin{itemize}[leftmargin=*]

\item \textbf{RQ1}: What is the performance and generalization capability of \model\ in diverse zero-shot spatio-temporal prediction tasks?

\item \textbf{RQ2}: How does \model\ perform when compared to existing spatio-temporal models in classical supervised scenarios?

\item \textbf{RQ3}: What specific contributions do the proposed key components bring to enhance the capabilities of our \model\ model?

\item \textbf{RQ4}: Can the proposed model robustly handle the forecasting scenarios with varying spatio-temporal patterns?

\end{itemize}

\begin{table*}[t]
\renewcommand\arraystretch{1}
    \centering
    \small
    \caption{Our model's performance in zero-shot prediction is evaluated on three diverse datasets: NYC-taxi, NYC-bike, and NYC-crime, providing a comprehensive assessment of its predictive capabilities in unseen situations.}
    \vspace{-0.15in}
    \label{tab:main_cross-region}
    \scalebox{1.07}{
    \begin{tabular}{c c |c c c c| c c c c| c c c c}
        \hline
        \multirow{3}*{Model} & \multicolumn{1}{c}{Dataset} & \multicolumn{4}{c}{NYC-taxi} & \multicolumn{4}{c}{NYC-bike} & \multicolumn{4}{c}{NYC-crime} \\
        \cline{2-14}
        & \multicolumn{1}{c}{Type} & \multicolumn{2}{c}{Inflow} & \multicolumn{2}{c}{Outflow} & \multicolumn{2}{c}{Inflow} & \multicolumn{2}{c}{Outflow} & \multicolumn{2}{c}{Burglary} & \multicolumn{2}{c}{Robbery}\\
        \cline{2-14}
        & \multicolumn{1}{c}{Metrics} & \multicolumn{1}{c}{MAE} & \multicolumn{1}{c}{RMSE} & \multicolumn{1}{c}{MAE} & \multicolumn{1}{c}{RMSE} & \multicolumn{1}{c}{MAE} & \multicolumn{1}{c}{RMSE} & \multicolumn{1}{c}{MAE} & \multicolumn{1}{c}{RMSE} & \multicolumn{1}{c}{Macro-F1} & \multicolumn{1}{c}{Recall} & \multicolumn{1}{c}{Macro-F1} & \multicolumn{1}{c}{Recall}\\
        \cline{1-14}
        \multicolumn{2}{c}{AGCRN} & 10.86 & 26.51 & 13.15 & 36.45 & 3.41 & 7.98 & 3.42 & 8.08 & 0.48 & 0.00 & 0.49 & 0.01 \\
        \multicolumn{2}{c}{ASTGCN} & 9.75 & 24.12 & 12.42 & 33.28 & 5.58 & 11.58 & 5.78 & 12.29 & 0.49 & 0.01 & 0.55 & 0.09 \\
        \multicolumn{2}{c}{GWN} & 10.73 & 26.50 & 9.67 & 26.74 & 3.32 & 8.17 & 3.07 & 7.52 & 0.48 & 0.00 & 0.52 & 0.04 \\
        \multicolumn{2}{c}{MTGNN} & 10.16 & 25.84 & 12.59 & 35.38 & 3.18 & 7.62 & 3.20 & 7.65 & 0.64 & 0.27 & 0.65 & 0.30 \\
        \multicolumn{2}{c}{STWA} & 11.28 & 28.97 & 13.54 & 38.61 & 4.59 & 10.94 & 4.35 & 10.67 & 0.48 & 0.00 & 0.51 & 0.03 \\
        \multicolumn{2}{c}{STSGCN} & 18.97 & 41.38 & 20.07 & 45.79 & 6.85 & 14.98 & 6.54 & 14.77 & 0.48 & 0.00 & 0.48 & 0.00 \\
        \multicolumn{2}{c}{STGCN} & 12.54 & 30.80 & 14.32 & 39.58 & 4.11 & 9.21 & 4.45 & 9.62 & 0.48 & 0.00 & 0.64 & 0.30 \\
        \multicolumn{2}{c}{TGCN} & 10.04 & 25.10 & 10.98 & 30.03 & 2.88 & 6.55 & 2.91 & 6.42 & 0.56 & 0.10 & 0.58 & 0.13 \\
        \multicolumn{2}{c}{DMVSTNET} & 11.00 & 28.29 & 10.59 & 29.20 & 3.80 & 9.87 & 3.65 & 9.21 & 0.48 & 0.01 & 0.59 & 0.15 \\
        \multicolumn{2}{c}{ST-LSTM} & 16.97 & 34.43 & 18.93 & 44.10 & 7.78 & 15.41 & 6.92 & 17.12 & 0.48 & 0.00 & 0.49 & 0.03 \\
        \hline
        \multicolumn{2}{c}{\model} & \textbf{6.16} & \textbf{16.92} & \textbf{6.83} & \textbf{21.78} & \textbf{2.02} & \textbf{5.16} & \textbf{2.01} & \textbf{5.03} & \textbf{0.67} & \textbf{0.34} & \textbf{0.69} & \textbf{0.42} \\
        \cline{1-14}
    \end{tabular}
    }
    \vspace{-0.05in}
\end{table*}

\subsection{Experimental Setting}
\subsubsection{\bf Dataset Description} To evaluate the effectiveness of the proposed model in predicting spatio-temporal patterns across various urban computing scenarios, we conducted experiments using four distinct datasets: NYC-taxi, NYC-bike, NYC-crime, and CHI-taxi. These datasets encompass a wide range of data sources to capture the dynamic nature of urban environments, including records of taxi travel, bike trajectories, crime incidents in New York City, and taxi travel data in Chicago. To facilitate our analysis, we partitioned the cities into grid-like regions based on latitude and longitude information. Within specific time intervals, we aggregated statistical measures for each region. For example, this involved calculating the number of taxi inflows and outflows within a 30-minute period in region A, or determining the count of theft incidents within a day in region B. Furthermore, Points of Interest (POIs) data can be obtained through APIs provided by map services, utilizing the latitude and longitude of different regions. For more comprehensive data descriptions, please refer to the Appendix.

\subsubsection{\bf Evaluation Protocols} 
In order to investigate the capabilities of large language models in analyzing diverse spatio-temporal data across different regions, we selected a subset of taxi, bike, and crime data from various areas of New York City as our training set. 
\begin{itemize}[leftmargin=*]

\item \textbf{Zero-Shot Learning Scenarios}. We assessed the model performance by predicting future spatio-temporal data from regions in NYC or even Chicago that were unseen in the training phase.

\item \textbf{Supervised Learning Scenarios}. We evaluated the model using future data from the same regions as the training set.

\end{itemize}
For regression tasks, we maintained a consistent training and testing methodology across all baseline models. When it came to classification tasks involving crime data, we utilized binary cross-entropy as the loss function for training and testing the models. Our experiments were conducted using the robust vicuna-7b~\cite{zheng2023judging} as the foundational large language model for UrbanGPT. For a more comprehensive understanding of our methodology and experimental setup, please refer to the appendix for detailed information.

\subsubsection{\bf Evaluation Metrics}
For regression tasks, we employed MAE (Mean Absolute Error) and RMSE (Root Mean Square Error) as evaluation metrics. These metrics quantify the discrepancies between the predicted outcomes and the actual labels, with lower values indicating superior performance~\cite{zheng2020gman,jiang2023pdformer}. In the case of classification tasks, we utilized Recall and Macro-F1 as evaluation metrics to assess performance. Recall measures the model's ability to correctly identify positive instances, while Macro-F1 is a comprehensive performance metric that combines precision and recall to provide an overall measure of classification accuracy~\cite{huang2018deepcrime,wang2021gsnet}.

\subsubsection{\bf Baseline Model}
We conducted a thorough comparison with 10 advanced models to establish baselines for our proposed method. \textbf{(i)} In the category of RNNs-based spatio-temporal forecasting methods, we compared our proposed method with AGCRN~\cite{bai2020adaptive}, DMVSTNET~\cite{yao2018deep} and ST-LSTM~\cite{libcity}. These approaches leverage RNNs for modeling and prediction. \textbf{(ii)} The GNNs-based spatio-temporal models primarily utilize graph neural networks to capture spatial correlations and integrate temporal encoders to capture spatio-temporal relationships. The models we compared against in this category include GWN~\cite{wu2019graph}, MTGNN~\cite{wu2020connecting}, STSGCN~\cite{Song2020spatial}, TGCN~\cite{Zhao2020TGCN}, and STGCN~\cite{yu2018spatio}. \textbf{(iii)} In the attention-based spatio-temporal models category, the methods employ attention mechanisms to model spatio-temporal correlations. The models we compared against in this category are ASTGCN~\cite{Guo2019attention} and STWA~\cite{cirstea2022towards}.

\subsection{Zero-Shot Prediction Performance (RQ1)}
\label{sec:mainresult}
In this section, we thoroughly evaluate the predictive performance of our proposed model in zero-shot scenarios. The results of our evaluation are presented in Table~\ref{tab:main_cross-region} and visualized in Figure~\ref{fig:cross-city}. Our objective is to assess the model's effectiveness in predicting spatio-temporal patterns in geographical areas that it has not encountered during training. This evaluation encompasses both cross-region and cross-city settings, allowing us to gain insights into the model's generalization capabilities across different locations.

\subsubsection{\bf Prediction on Unseen Regions within a City}
Cross-region scenarios entail using data from certain regions within a city to forecast future conditions in other regions that have not been encountered by the model. Through a thorough analysis of the model's performance in these cross-region predictions, we can draw attention to three significant observations: \\\vspace{-0.12in}

\noindent \textbf{i) Superior Zero-shot Predictive Performance}. The results presented in Table~\ref{tab:main_cross-region} highlight the exceptional performance of our proposed model in both regression and classification tasks on various datasets, surpassing the baseline models in zero-shot prediction. The success of our model can be attributed to two key factors.
\begin{itemize}[leftmargin=*]

\item \textbf{Spatio-Temporal-Text-Alignment}. The alignment of spatio-temporal contextual signals with the text comprehension abilities of language models plays a pivotal role in the success of our proposed model. This fusion enables the model to effectively capitalize on both the encoded urban dynamics from the spatiotemporal signals and the comprehensive understanding of textual context provided by the LLMs. By leveraging these two essential aspects, our model achieves the remarkable ability to generalize its prediction capabilities in zero-shot scenarios.

\item \textbf{Spatio-Temporal Instruction-Tuning}. This adaptive tuning process empowers the LLM to effectively integrate crucial information from the instructions, enhancing its comprehension of the complex relationships and dependencies between spatial and temporal factors. By seamlessly merging the spatio-temporal instruction-tuning with the spatio-temporal dependency encoder, our proposed model, \model, successfully preserves universal and transferable spatio-temporal knowledge. Consequently, the model becomes capable of capturing the fundamental patterns and dynamics that govern spatio-temporal phenomena, enabling it to make precise predictions in downstream zero-shot scenarios.

\end{itemize}

\noindent \textbf{ii) Enhanced Urban Semantic Understanding}. Urban semantics offer valuable insights into the diverse dimensions of spatial and temporal characteristics. Our approach involves training our model on a wide range of datasets, enriching its understanding of spatio-temporal dynamics across different timeframes and geographical locations. In contrast, baseline models tend to prioritize encoding temporal and spatial dependencies, neglecting the nuanced semantics that differentiate regions, timeframes, and data categories. By incorporating comprehensive semantic awareness into our \model, we significantly enhance its ability to make accurate zero-shot predictions in previously unseen regions. \\\vspace{-0.12in}

\noindent \textbf{iii) Improved Performance in Sparse Data Scenarios}. Predicting spatio-temporal patterns in sparse data environments is challenging as models tend to overfit when data points are scarce. This challenge is particularly notable when predicting crimes, where data is often sparse but crucial for accurate predictions. Baseline models struggle in cross-regional prediction tasks under these sparse conditions, resulting in low recall scores that indicate potential overfitting. To overcome this limitation, our model integrates spatio-temporal learning with large language models (LLMs) using an effective spatio-temporal instruction-tuning paradigm. By incorporating rich semantic insights, our approach enhances the model's spatio-temporal representations, enabling it to effectively handle sparse data and achieve improved prediction accuracy.

\subsubsection{\bf Cross-City Prediction Task.}
\begin{figure*}
\centering
\subfigure{
    \begin{minipage}[t]{0.25\linewidth}
        \centering
        \includegraphics[width=1.5in]{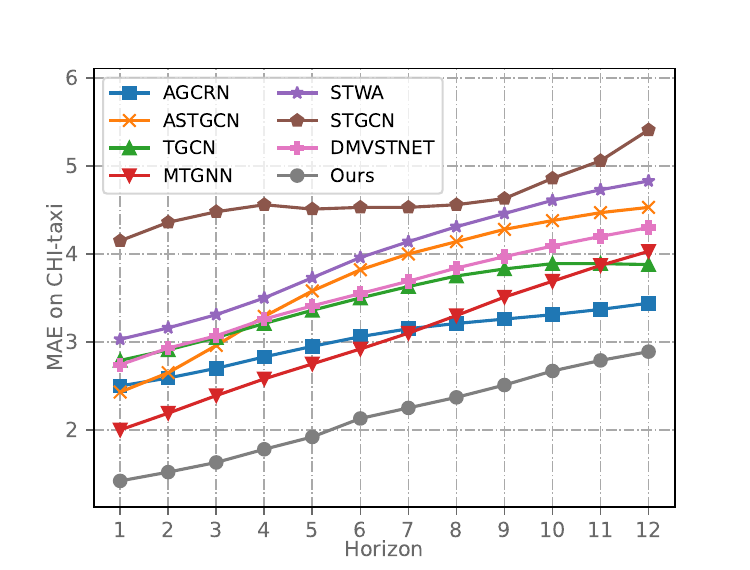}\\
        \vspace{-1cm}
    \end{minipage}%
}%
\subfigure{
    \begin{minipage}[t]{0.25\linewidth}
        \centering
        \includegraphics[width=1.5in]{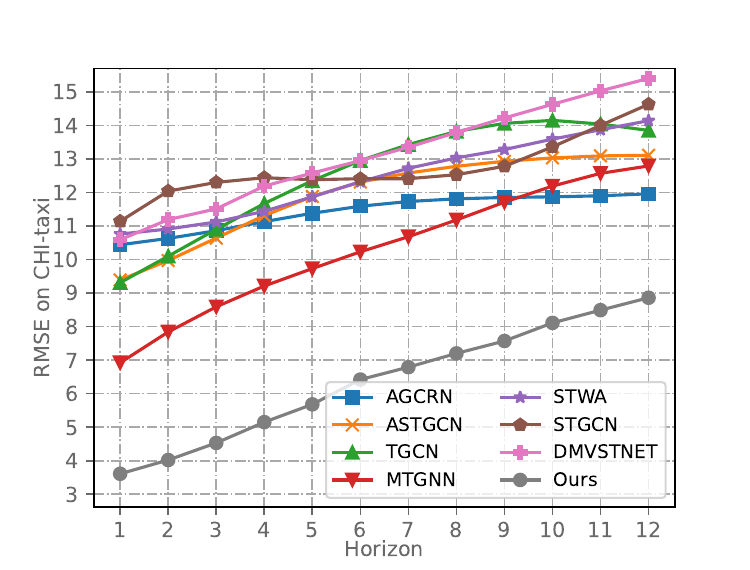}\\
        \vspace{-1cm}
    \end{minipage}%
}%
\subfigure{
    \begin{minipage}[t]{0.25\linewidth}
        \centering
        \includegraphics[width=1.5in]{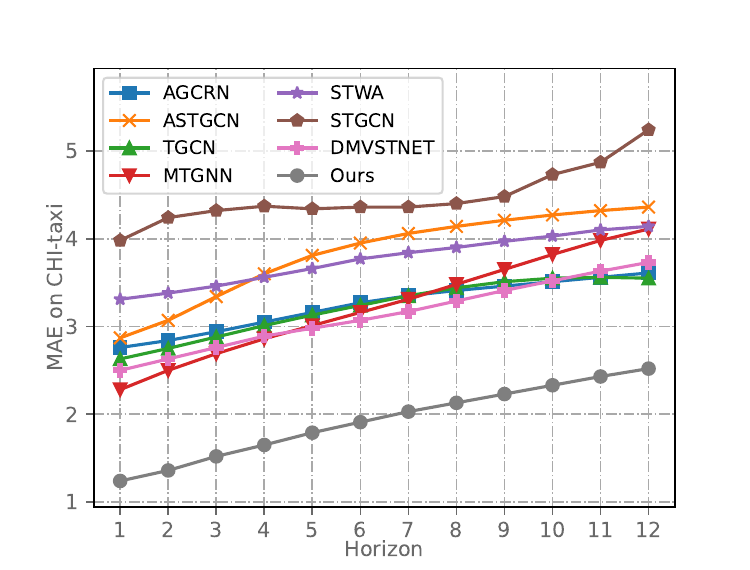}\\
        \vspace{-1cm}
    \end{minipage}%
}%
\subfigure{
    \begin{minipage}[t]{0.25\linewidth}
        \centering
        \includegraphics[width=1.5in]{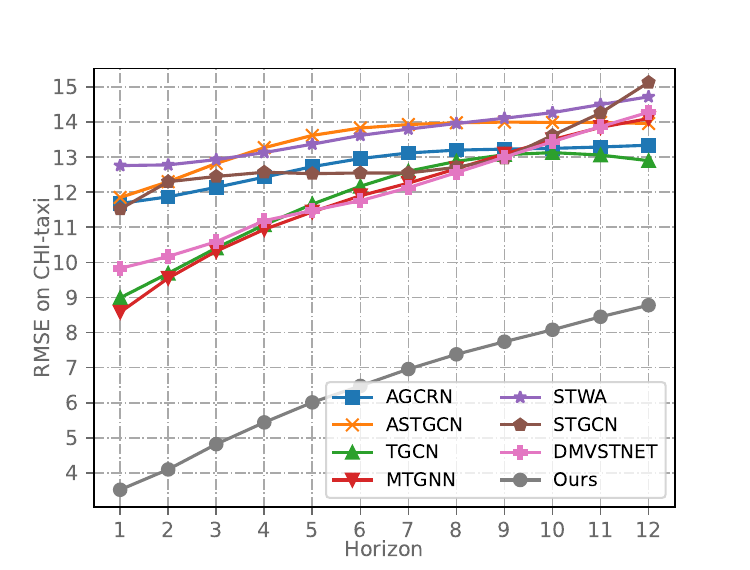}\\
        \vspace{-1cm}
    \end{minipage}%
}%
\centering
\vspace{-0.4cm}
\caption{Time step-based prediction comparison experiment conducted on the CHI-taxi dataset. }
\vspace{-0.2cm}
\label{fig:cross-city}
\end{figure*}

To assess the performance of our model in cross-city prediction tasks, we conducted tests on the CHI-taxi dataset, which was not seen during the training phase. The results, depicted in Figure~\ref{fig:cross-city}, yielded the following observations:
\begin{itemize}[leftmargin=*]

\item \textbf{Consistency in Multi-step Prediction}: Our model consistently outperforms the comparison method at each time step. Notably, it maintains a significant advantage in both short-term and long-term spatio-temporal prediction, demonstrating the robustness of our proposed model in cross-city prediction scenarios.

\item \textbf{Effective Knowledge Transfer Across Cities}: The prediction results obtained from the CHI-taxi dataset validate the superior forecasting capabilities of our model in cross-city scenarios. This enhancement can be attributed to the integration of spatio-temporal encoders with the spatio-temporal instruction-tuning paradigm. By incorporating these components, our model effectively captures universal and transferable spatio-temporal patterns, allowing it to make accurate predictions. Additionally, by considering different geographical information and temporal factors alongside the learned transferred knowledge, our model successfully associates spatio-temporal patterns exhibited by similar functional areas and historical periods. This comprehensive understanding provides valuable insights for making precise zero-shot predictions in cross-city scenarios.

\end{itemize}

\vspace{-0.1in}
\subsection{Classical Supervised Prediction Task (RQ2)}
This section examines the predictive capabilities of our \model\ in end-to-end supervised prediction scenarios, as presented in Table~\ref{tab:cross-temporal}. We will discuss the results from two perspectives below. \\\vspace{-0.12in}

\noindent \textbf{Enhanced Long-Term Forecasting Abilities}: We examined the model's effectiveness in long-term spatio-temporal forecasting by utilizing test datasets that spanned broader time intervals. For instance, we trained the model using data from 2017 and evaluated its performance on data from 2021. The results of these experiments demonstrate that our \model\ possesses a significant advantage over the baselines, highlighting its superior ability to generalize across different temporal landscapes. This capability reduces the need for frequent retraining or incremental updates, making the model more aligned with real-world applications. Additionally, the experiments have confirmed that incorporating additional textual knowledge does not hinder model performance or introduce noise, thus further validating the feasibility of utilizing large language models to enhance spatio-temporal forecasting tasks. \\\vspace{-0.12in}

\noindent \textbf{Spatial Semantic Understanding}: Accurately capturing spatial correlations is crucial in the realm of spatio-temporal prediction. Traditional methods often employ graph networks or attention mechanisms to analyze these correlations. Models lacking dedicated spatial correlation modules, like LSTM, tend to underperform as they overlook the spatial context. In contrast, our model compensates for the absence of explicit spatial encoders by integrating extensive geographic and points of interest (POIs) data within the textual input. This approach enables the model to comprehend the shared characteristics of areas with similar functions at a higher semantic level. Consequently, it deduces patterns of correlation between various functional zones and effectively represents the interconnections among different regions.

\begin{table}[t]
\renewcommand\arraystretch{1}
    \centering
    \caption{Evaluation of performance in the end-to-end supervised setting on the NYC-taxi and NYC-bike datasets.}
    \vspace{-0.15in}
    \label{tab:cross-temporal}
    \scalebox{0.8}{
    \begin{tabular}{l|c c|c c|c c|c c}
        \hline
        \multirow{3}{*}{Model} & \multicolumn{4}{c|}{NYC-taxi} & \multicolumn{4}{c}{NYC-bike}\\
        \cline{2-9}
        & \multicolumn{2}{c|}{Inflow} & \multicolumn{2}{c|}{Outflow} & \multicolumn{2}{c|}{Inflow} & \multicolumn{2}{c}{Outflow} \\
        \cline{2-9}
        & \multicolumn{1}{c}{MAE} & \multicolumn{1}{c|}{RMSE} & \multicolumn{1}{c}{MAE} & \multicolumn{1}{c|}{RMSE} & \multicolumn{1}{c}{MAE} & \multicolumn{1}{c|}{RMSE} & \multicolumn{1}{c}{MAE} & \multicolumn{1}{c}{RMSE}\\
        \hline
        AGCRN & 2.83 & 8.35 & 2.62 & 9.21 & 3.30 & 7.65 & 3.38 & 7.73 \\
        ASTGCN & 5.41 & 18.04 & 5.00 & 19.29 & 3.87 & 7.93 & 3.66 & 7.69 \\
        GWN & 3.91 & 11.93 & 2.89 & 10.85 & 4.30 & 9.04 & 3.88 & 8.29 \\
        MTGNN & 3.09 & 10.13 & 2.61 & 10.96 & 3.31 & 7.47 & 3.26 & 7.61 \\
        STWA & 3.90 & 12.64 & 3.15 & 11.32 & 4.23 & 9.07 & 4.18 & 9.18 \\
        STSGCN & 4.57 & 13.93 & 4.41 & 15.87 & 5.10 & 12.23 & 4.72 & 10.78 \\
        STGCN &3.45 & 9.82 & 3.17 & 10.53 & 3.88 & 9.23 & 3.90 & 9.08 \\
        TGCN & 3.99 & 11.47 & 3.31 & 11.58 & 4.12 & 7.92 & 4.11 & 7.84 \\
        DMVSTNET & 3.83 & 11.55 & 2.76 & 9.88 & 3.71 & 7.95 & 3.69 & 7.92 \\
        ST-LSTM & 7.78 & 15.41 & 6.92 & 17.12 & 5.00 & 11.52 & 4.96 & 11.41\\
        \hline
        \emph{\model} & \textbf{2.50} & \textbf{6.78} & \textbf{1.71} & \textbf{6.68} & \textbf{3.11} & \textbf{7.10} & \textbf{3.01} & \textbf{6.94}\\
        \hline
    \end{tabular}
    \vspace{-0.05in}
    }
\end{table}

\subsection{Ablation study (RQ3)}
\begin{figure}
\centering
\subfigure{
    \begin{minipage}[t]{1\linewidth}
        \centering
        \includegraphics[width=2.3in]{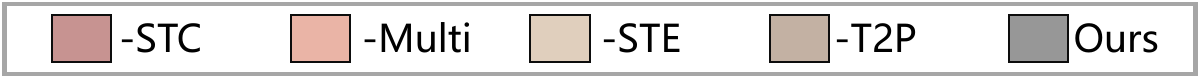}
    \end{minipage}%
}%
\vspace{-0.1in}
\subfigure{
    \vspace{-0.08in}
    \begin{minipage}[t]{0.25\linewidth}
        \centering
        \includegraphics[width=0.8in]{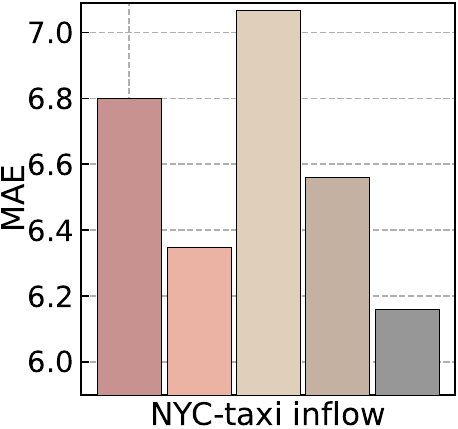}
    \end{minipage}%
    \begin{minipage}[t]{0.25\linewidth}
        \centering
        \includegraphics[width=0.83in]{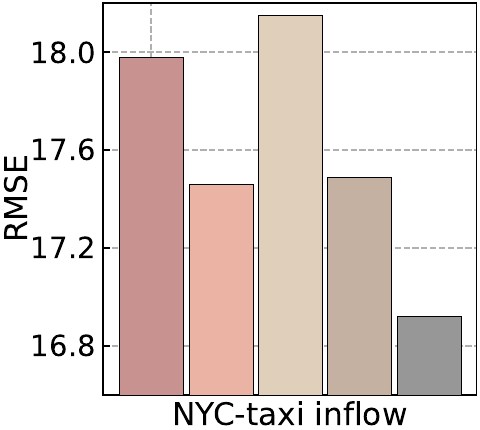}
    \end{minipage}%
    \begin{minipage}[t]{0.25\linewidth}
        \centering
        \includegraphics[width=0.8in]{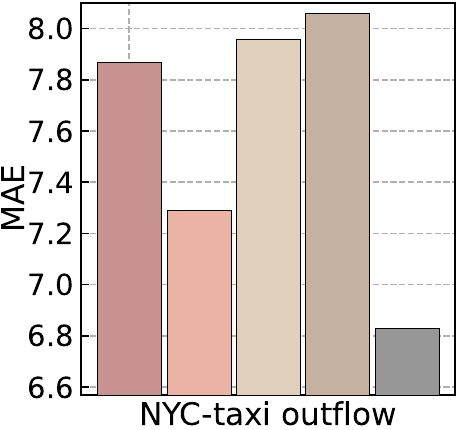}
    \end{minipage}%
    \begin{minipage}[t]{0.25\linewidth}
        \centering
        \includegraphics[width=0.83in]{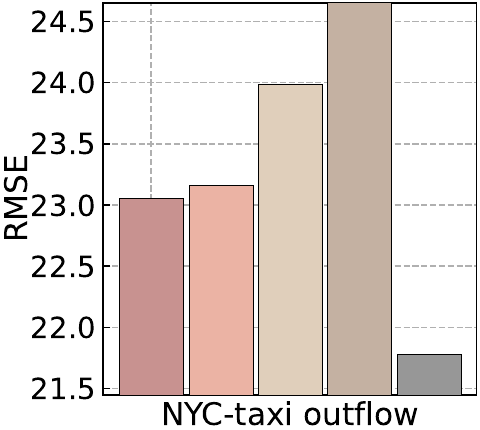}
    \end{minipage}%
}%
\centering
\vspace{-0.4cm}
\caption{Ablation study of our proposed \model.}
\vspace{-0.2in}
\label{fig:ablation}
\end{figure}

This section investigates the impact of different key components on the performance of our model, as illustrated in Figure~\ref{fig:ablation}. Our rigorous testing primarily revolves around the zero-shot scenario using the NYC-taxi dataset. Through our analysis, we have distilled the benefits offered by the different modules into four key points.

\begin{itemize}[leftmargin=*]

\item \textbf{(1) Impact of Spatial and Temporal Context}: \textbf{-STC}. By removing time and spatial information from the instruction text, we observed a noticeable decline in the model's performance.

\end{itemize}

This can be attributed to the lack of temporal information, forcing the model to rely solely on spatio-temporal encoders for encoding time-related features and performing prediction tasks. Furthermore, the absence of spatial information hindered the model's ability to capture spatial correlations, making it challenging to analyze the distinct spatio-temporal patterns of different areas.


\begin{itemize}[leftmargin=*]

\item \textbf{(2) Impact of Instruction-Tuning with Diverse Datasets}:
\textbf{-Multi}. We conducted our training solely on the NYC-taxi data to examine whether incorporating multiple datasets would provide valuable insights to the LLMs in zero-shot scenarios. 

\end{itemize}

However, this restricted training approach limited the model's ability to fully uncover the spatio-temporal dynamics of cities due to the absence of diverse urban indicators. As a result, the model's performance was suboptimal. By integrating diverse spatio-temporal data from multiple sources, our model can effectively capture the unique characteristics of different geographical locations and their evolving spatio-temporal patterns.

\begin{itemize}[leftmargin=*]

\item \textbf{(3) Impact of Spatio-Temporal Encoder}:
\textbf{-STE}. In this variant, we disable the spatio-temporal encoder to investigate its effect on aligning the large language model with encoded urban dependency dynamics into latent embedding space.

\end{itemize}

The results clearly indicate that the absence of the spatio-temporal encoder significantly hampers the performance of the large language model in spatio-temporal prediction scenarios. This underscores the crucial role played by the proposed spatio-temporal encoder in enhancing the model's predictive capabilities.

\begin{itemize}[leftmargin=*]

\item \textbf{(4) Regression Layer Incorporation in Instruction-Tuning:} \textbf{T2P}. We explicitly instructed \model\ to generate its predictions in a textual format. However, the suboptimal performance indicates the challenges in utilizing LLMs for precise numerical regression tasks, as opposed to employing regression layers.

\end{itemize}

The primary challenge stems from the model's dependence on multi-class loss for optimization during training, leading to a mismatch between the model's probabilistic output and the continuous value distribution necessary for spatio-temporal forecasting. To bridge this gap, we incorporated a regression predictor into our model, significantly improving its capacity to generate more precise numerical predictions for regression tasks.

\subsection{Model Robustness Study (RQ4)}
\begin{figure}
\centering
\subfigure{
    \begin{minipage}[t]{1\linewidth}
        \centering
        \includegraphics[width=3.0in]{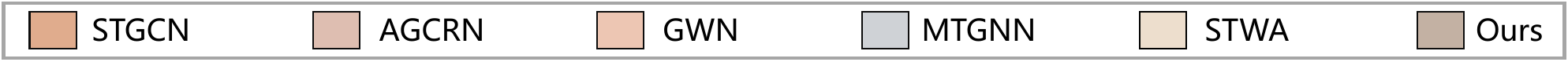}
    \end{minipage}%
}%
\\
\vspace{-0.1in}
\subfigure{
    \vspace{-0.08in}
    \begin{minipage}[t]{0.5\linewidth}
        \centering
        \includegraphics[width=1.45in]{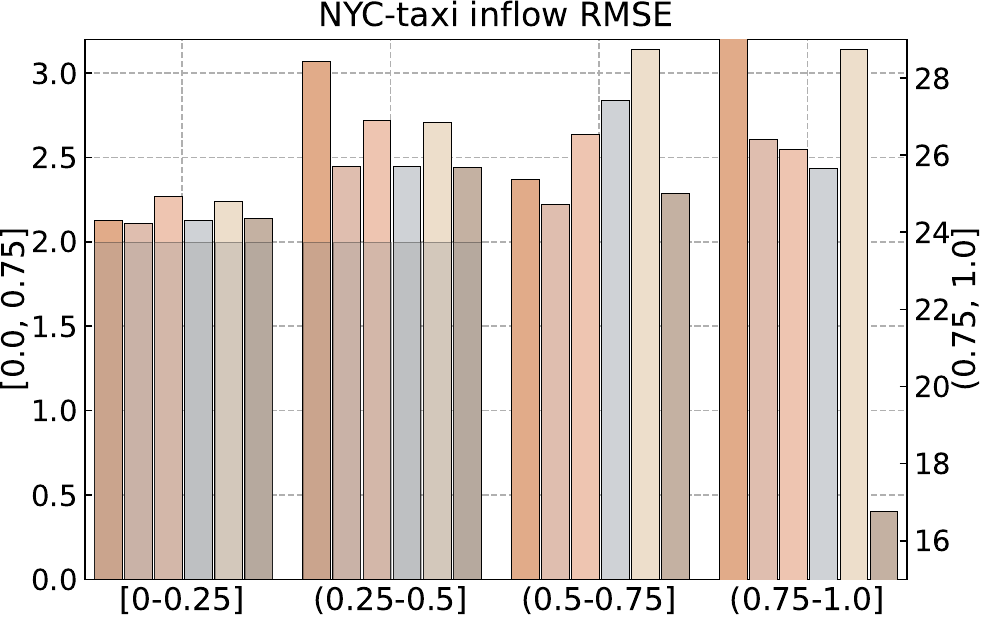}
    \end{minipage}%
    \begin{minipage}[t]{0.5\linewidth}
        \centering
        \includegraphics[width=1.45in]{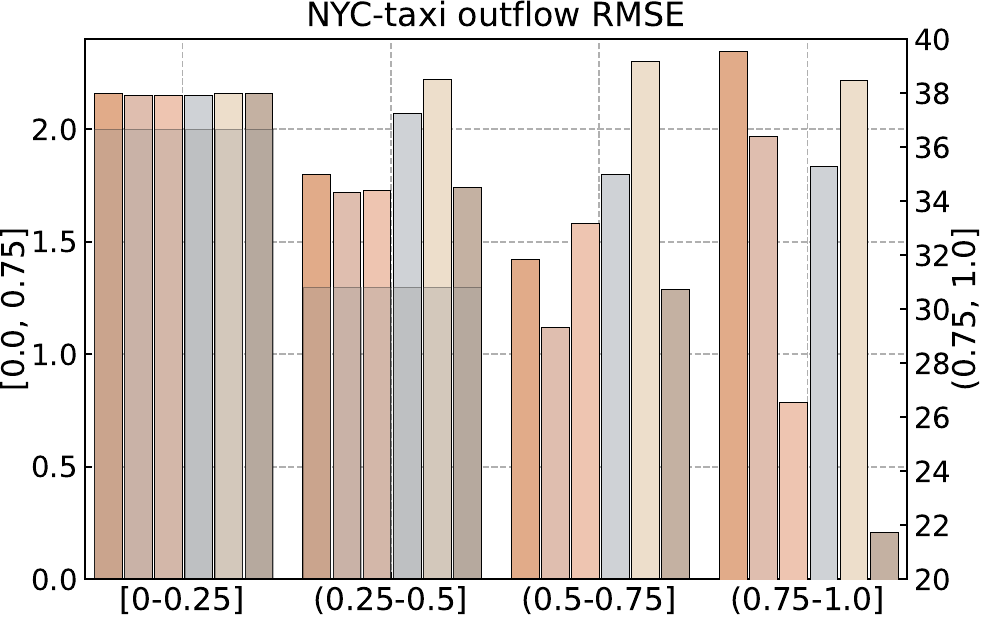}
    \end{minipage}%
}%
\\
\subfigure{
    \vspace{-0.08in}
    \begin{minipage}[t]{0.5\linewidth}
        \centering
        \includegraphics[width=1.45in]{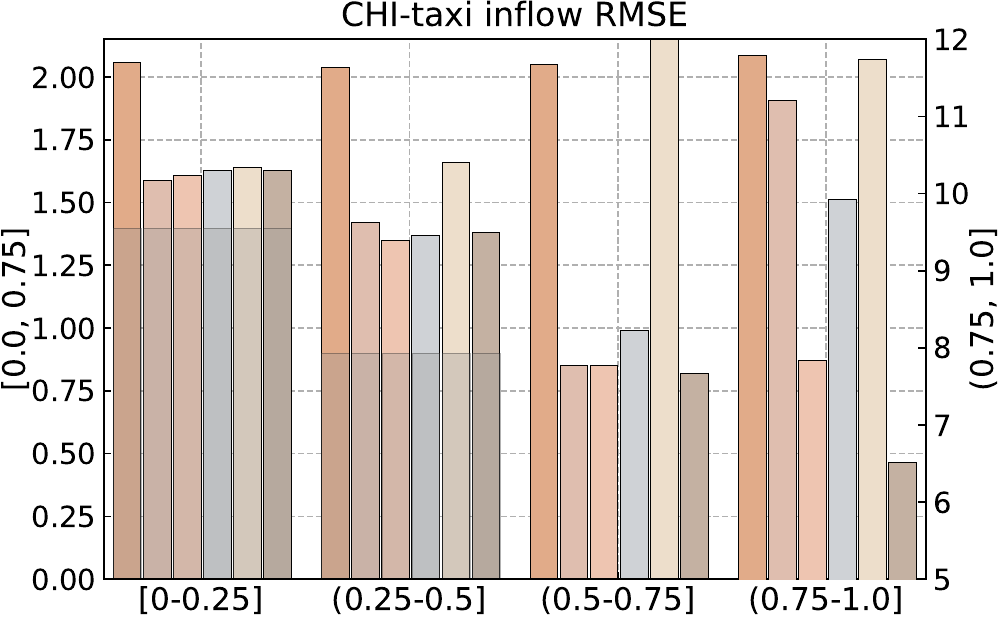}
    \end{minipage}%
    \begin{minipage}[t]{0.5\linewidth}
        \centering
        \includegraphics[width=1.45in]{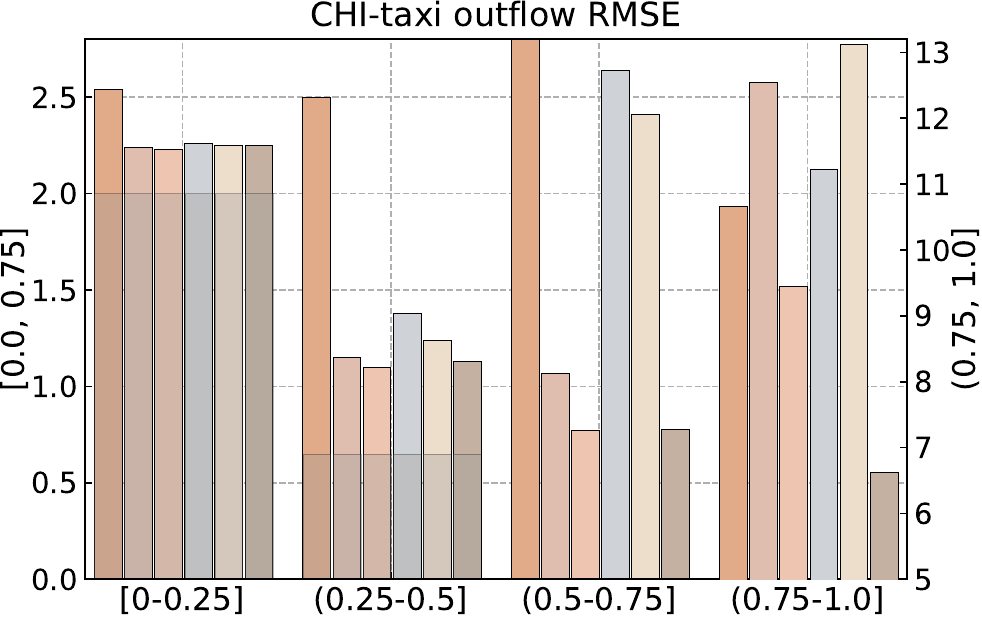}
    \end{minipage}%
}%
\centering
\vspace{-0.4cm}
\caption{Robustness study of the \model\ model.}
\vspace{-0.25in}
\label{fig:robust}
\end{figure}

In this section, we focus on evaluating the robustness of our \model\ across different spatio-temporal pattern scenarios. We categorize regions based on the magnitude of numerical variations, such as taxi flow, during a specific time period. Lower variance indicates stable temporal patterns, while higher variance suggests diverse spatio-temporal patterns in active commercial zones or densely populated areas. Our findings, shown in Figure~\ref{fig:robust}, reveal that most models perform well in regions with lower variance, where patterns remain relatively stable. However, the baseline model struggles in regions with high variance, particularly within the (0.75, 1.0] range, resulting in inaccurate predictions. This limitation may stem from the baseline model's difficulty in inferring spatio-temporal patterns in unseen regions during zero-shot scenarios. In practical applications, accurate prediction of densely populated or bustling areas is crucial for urban governance, such as traffic light control and security scheduling. Our \model\ demonstrates significant performance improvement in the (0.75, 1.0] interval, highlighting its powerful zero-shot prediction capability with our method.

\vspace{-0.05in}
\subsection{Case Study}
In our case study, we thoroughly evaluate several large language models (LLMs) for zero-shot spatio-temporal prediction. We emphasize the challenges these models face in directly understanding spatio-temporal patterns from numeric geo-series data. In contrast, we showcase the exceptional performance of our proposed \model\ framework in capturing universal spatio-temporal patterns and its ability to generalize effectively across various zero-shot spatio-temporal forecasting scenarios. For a more comprehensive understanding, please refer to the Appendix.

\vspace{-0.05in}
\section{Related Work}
\label{sec:relate}

\noindent \textbf{Deep Spatio-temporal Prediction Models}. 
Deep spatio-temporal prediction methods have gained prominence in deep learning due to their impressive performance. These models typically consist of two components: temporal dependency modeling and spatial correlation encoding. Early models like D-LSTM~\cite{yu2017deep} and ST-resnet~\cite{zhang2017deep} used RNNs and convolutional networks to model temporal and spatial dependencies. Graph neural networks (GNNs) proved to be a natural fit for spatial correlation modeling, as seen in models like STGCN~\cite{yu2018spatio} and DCRNN~\cite{li2017diffusion}, which utilized graph structures based on node distances. Techniques such as learnable region-wise graph structures~\cite{wu2019graph,wu2020connecting} and dynamic spatio-temporal graph networks~\cite{han2021dynamic,zhao2023dynamic} have further enhanced spatial correlation modeling.

Furthermore, researchers have explored approaches such as multi-scale temporal learning~\cite{wang2020traffic} and multi-granularity temporal learning~\cite{Guo2019attention} to encode temporal dependencies. These strategies enable the capture of features like long-term and short-term correlations as well as periodicity. These advancements have significantly contributed to the progress of spatio-temporal prediction. However, it is worth noting that the majority of these studies are tailored for supervised contexts, with limited research and development focused on zero-shot spatio-temporal forecasting. This represents an important area that requires further exploration. \\\vspace{-0.12in}

\noindent \textbf{Spatio-Temporal Pre-training}. 
Spatio-temporal pre-training techniques have recently received significant research attention. These techniques primarily focus on generative~\cite{li2023gpt,shao2022pretraining} and contrastive~\cite{zhang2023automated} pre-training models to enhance the predictive performance of downstream tasks. There has also been extensive exploration of pretrain-finetune frameworks for few-shot learning scenarios~\cite{lu2022spatio,jin2022selective}, aiming to improve knowledge transferability by aligning source and target data. However, these approaches require training or fine-tuning on target data and lack zero-shot prediction capabilities. In this work, we address the challenge of data scarcity in downstream urban scenarios by proposing \model. Our model demonstrates the ability to generalize well across various scenarios, mitigating the need for extensive training or fine-tuning on target data. \\\vspace{-0.12in}

\noindent \textbf{Large Language Models.} 
The emergence of large language models~\cite{brown2020language, ouyang2022training} has recently attracted significant attention due to their unprecedented machine performance in tasks like text understanding and reasoning. These models have become a hot topic, demonstrating the potential to advance from intelligent algorithms to artificial intelligence. Open-source large language models such as Llama~\cite{touvron2023llama,touvron2023llama2}, Vicuna~\cite{zheng2023judging}, and ChatGLM~\cite{zeng2023glm} have been released, leading to researchers exploring their application in various fields to enhance transfer learning capabilities with domain knowledge from these models. In the computer vision domain, researchers have combined multimodal large language models with prompt learning methods to achieve zero-shot predictions in downstream tasks~\cite{shen2024multitask,zhou2022learning,zhu2023minigpt4}. Furthermore, the capabilities of LLMs in graph reasoning~\cite{besta2024graph,chai2023graphllm,tang2023graphgpt}, recommendation~\cite{wei2023llmrec,harte2023leveraging,ren2023representation} and traffic analysis~\cite{da2023open} have been extensively studied. However, the utilization of large language models for zero-shot spatio-temporal prediction tasks in the field of urban intelligence remains largely unexplored.
\vspace{-0.05in}
\section{Conclusion}
\label{sec:conclusoin}

We present the \model, a spatio-temporal large language model, to generalize well in diverse urban scenarios. To seamlessly align the spatio-temporal contextual signals with LLMs, we introduce a spatio-temporal instruction-tuning paradigm. This empowers the \model\ with the remarkable ability to learn universal and transferable spatio-temporal patterns across various types of urban data. Through extensive experiments and meticulous ablation studies, we demonstrate the exceptional effectiveness of the \model's architecture and its key components. However, it is important to acknowledge that while the results are promising, there are still limitations to be addressed in future studies. As a first step, we are actively engaged in collecting a more diverse range of urban data to enhance and refine the capabilities of our \model\ across a broader spectrum of urban computing domains. Additionally, understanding the decision-making process of our \model\ is of importance. While the model demonstrates exceptional performance, providing interpretability and explainability is equally essential. Future research efforts will focus on empowering our model with the ability to interpret and explain its predictions.

\clearpage

\bibliographystyle{ACM-Reference-Format}
\balance
\bibliography{refers} 

\clearpage
\appendix \section{Appendix}
\balance
\label{sec:appendix}
In the appendix section, we offer comprehensive discussion of the experimental setup. This includes detailed dataset information, hyperparameter configuration, experimental setting for the instruction-tuning phase and test phase, as well as the baselines. Furthermore, we present a case study of our \model, showcasing its effectiveness for zero-shot spatio-temporal predictions.

\subsection{Experimental Details Description}
\subsubsection{\bf Dataset Details.} 
We collected data on taxi flows, bicycle flows, and crime incidents in New York City for training and evaluation. The NYC-taxi dataset contains 263 regions, with each region measuring approximately 3km x 3km. The time sampling interval for this dataset is 30 minutes. The NYC-bike and NYC-crime datasets consist of 2162 regions, each represented by a 1km x 1km grid. The time sampling interval for NYC-bike is also 30 minutes, while for NYC-crime, it is 1 day. All datasets cover the time period from Jan 1, 2016, to Dec 31, 2021, in NYC. The CHI-taxi dataset includes 77 regions, with each region measuring approximately 4km x 4km. This dataset includes all taxi data from January 1, 2021, to December 31, 2021, with a time sampling interval of 30 minutes.

\subsubsection{\bf Hyperparameters Settings}
The parameters for the dilation convolution kernel in the time encoder are set as follows: $d_{in}$, $d_{out}$, and $d_{out}^{'}$ are all set to 32, with a dilation factor of 1. For our prediction task, we aim to predict the next 12 steps of data based on the previous 12 steps. Both the history length ($H$) and prediction length ($P$) are set to 12. The projection layer parameters are configured with $d$ set to 64 and $d_L$ set to 4096. Lastly, the hidden layer parameter $d^{'}$ for the regression layer is set to 128.

\subsubsection{\bf Further Experimental Setup Descriptions} 
During the instruction-tuning phase, we randomly selected 80 regions from the three datasets in New York City as training data. It's important to note that the region indices were kept consistent for the NYC-bike and NYC-crime datasets. The training sets had specific time intervals: for NYC-taxi datasets, it ranged from January 1, 2017, to March 31, 2017; for NYC-bike datasets, it covered April 1, 2017, to June 30, 2017; and for NYC-crime datasets, it spanned from January 1, 2016, to December 31, 2018. For the pretraining of the spatio-temporal dependency encoder and the baseline models, we utilized the same data for training to optimize the parameters. The maximum epoch count was set at 100. In the testing phase, we conducted the following evaluations:
\begin{itemize}[leftmargin=*]

\item (i) \textbf{Zero-Shot Prediction}. 
We also selected an additional 80 regions from the New York City datasets as unseen test data. For the NYC-bike and NYC-taxi datasets, we used the first two weeks of data in 2020 for testing. As for the NYC-crime dataset, we used the entire year of 2020 for testing. In the case of the Chicago city dataset, we evaluated the model using all the data from Dec 2021.

\item (ii) \textbf{Classical Supervised Prediction}: For evaluation purposes, we chose the data with the longest time interval, which involved testing the model using all the data from the NYC-bike and NYC-taxi datasets specifically for the month of December 2021.

\end{itemize}

\subsubsection{\bf Baseline Details} 
To facilitate our discussion, we have categorized all the baselines into three distinct categories: RNN-based models, attention-based models, and graph neural network (GNN)-based spatio-temporal prediction models. Below, we provide detailed descriptions of each baseline category:

\noindent \textbf{RNN-based Spatio-Temporal Methods:}
\begin{itemize}[leftmargin=*]
\item \textbf{ST-LSTM}\cite{libcity}: It incorporates the Long Short-Term Memory to capture temporal dependencies in the spatio-temporal data.
\item \textbf{AGCRN}\cite{bai2020adaptive}: RNNs are employed to capture temporal correlations, allowing for the representation of the evolving patterns over time.
\item \textbf{DMVSTNET}\cite{yao2018deep}: 
In this method, RNNs are utilized to effectively model temporal dependencies, capturing the patterns that evolve over time. Furthermore, convolutional networks and fully connected layers are employed to capture local spatial correlations and establish meaningful spatial relationships.
\end{itemize}

\noindent \textbf{Attention-based Spatio-Temporal Approaches:}
\begin{itemize}[leftmargin=*]
\item \textbf{ASTGCN}\cite{Guo2019attention}: In this method, attention mechanisms are employed to capture multi-granularity temporal correlation features.
\item \textbf{STWA}\cite{cirstea2022towards}: The model incorporates personalized temporal and spatial parameters into the attention module, allowing for the modeling of dynamic spatio-temporal correlations.
\end{itemize}
\noindent \textbf{Spatio-Temporal GNNs:}
\begin{itemize}[leftmargin=*]
\item \textbf{GWN}\cite{wu2019graph}: 
It incorporates a learnable graph structure and 1-D convolutions to effectively learn spatio-temporal dependencies.
\item \textbf{MTGNN}\cite{wu2020connecting}: 
It utilizes a learnable graph structure to model multivariate temporal correlations. MTGNN employs 1-D dilation convolutions to generate temporal representations.
\item \textbf{TGCN}\cite{Zhao2020TGCN}: This model combines graph neural networks (GNNs) for spatial correlation modeling and recurrent neural networks (RNNs) for temporal correlation modeling.
\item \textbf{STGCN}\cite{yu2018spatio}: 
It uses gated temporal convolutions and GNNs to model temporal and spatial dependencies, respectively.
\item \textbf{STSGCN}\cite{Song2020spatial}: It introduces a spatio-temporal graph construction to learn spatial correlations across adjacent time steps.
\end{itemize}

\subsection{Case study}
In this section, we assess the effectiveness of different large language models (LLMs) in zero-shot spatio-temporal prediction scenarios, as illustrated in Table~\ref{tab:case1} and Table~\ref{tab:case2}. The instructions provided to the models are clearly indicated in blue font. The results demonstrate that various LLMs are capable of generating predictions based on these instructions, thereby highlighting the effectiveness of the prompt design. For instance, ChatGPT relies on historical averages rather than explicitly incorporating temporal or spatial data in its predictions. Llama-2-70b analyzes specific time periods and regions, but it encounters challenges in encoding numerical time-series dependencies, resulting in suboptimal predictive performance. On the other hand, Claude-2.1 effectively summarizes and analyzes historical data, leveraging peak-hour patterns and points of interest to achieve more accurate traffic trend predictions. 

Our proposed \model\ seamlessly integrates spatio-temporal contextual signals with the reasoning capabilities of large language models (LLMs) through a spatio-temporal instruction-tuning paradigm. This integration leads to remarkable improvements in predicting numerical values and spatio-temporal trends. These findings underscore the potential and effectiveness of our framework in capturing universal spatio-temporal patterns, making zero-shot spatio-temporal prediction practical and achievable.

\begin{table*}[htbp]
  \centering
  \caption{We examine the zero-shot predictions of different LLMs for bicycle flow in NYC with the provided instructions.}
  \vspace{-0.1cm}
  \begin{tabular}{p{17.3cm}}
    \toprule
    \textbf{Instructions:} Given the historical data for bike flow over 12 time steps in a specific region of New York City, the recorded bike inflows are [12 8 9 15 10 12 14 9 6 12 25 20], and the recorded bike outflows are [5 12 12 15 10 11 6 9 4 6 10 15]. The recording time of the historical data is 'January 14, 2020, 12:00, Tuesday to January 14, 2020, 17:30, Tuesday, with data points recorded at 30-minute intervals'. Here is the region information: This region is located within the Staten Island borough district and encompasses various POIs within a one-kilometer radius, covering Education Facility, Cultural Facility, Commercial, Transportation Facility categories. We now aim to predict the bike inflow and outflow for the next 12 time steps during the time period of 'January 14, 2020, 18:00, Tuesday to January 14, 2020, 23:30, Tuesday, with data points recorded at 30-minute intervals'. \textcolor{tblue}{To improve prediction accuracy, a spatio-temporal model is utilized to encode the historical bike data as tokens <ST\_Start><ST\_HIS><ST\_HIS><ST\_End>, where the first and the second tokens correspond to the representations of bike inflow and outflow.} Please conduct an analysis of the traffic patterns in this region, taking into account the provided time and regional information, and then generate the predictions (\textcolor{tblue}{the predictive tokens for regression}). \\
    \midrule
    \textbf{Ground true:} Based on the given information, the predictions of bike inflow and outflow in this region are \textbf{[26 38 28 24 20 13 16 9 5 3 5 2]} and \textbf{[12 19 10 10 11 9 6 5 4 4 2 2]}. \\
    \midrule
    \textbf{Llama-2-70b Responds:} Based on the provided historical data, I have ... Here are my predictions for the bike inflow and outflow for the next 12 time steps:Bike Inflow: \textbf{[15 18 20 15 12 10 14 12 10 15 18 20]}. Bike Outflow: \textbf{[10 12 15 10 11 12 10 9 6 10 12 15]}. \\
    These predictions are based on the following analysis: Time of Day: The traffic patterns during this time are likely to be influenced by the evening rush hour, with increased bike inflows and outflows during the peak hours...Regional Information: ... This suggests that there may be a mix of both commuter and recreational bike traffic in the area... \\
    \midrule
    \textbf{ChatGPT Responds:} To analyze the traffic patterns in the given region and make predictions for the next 12 time steps, we can calculate the average bike inflow and outflow based on the historical data... Now, we can generate predictions for the next 12 time steps: Predicted Bike Inflow: \textbf{[11.67 11.67... 11.67]}. Predicted Bike Outflow: \textbf{[9.58 9.58... 9.58]}...\\
    \midrule
    \textbf{Claude-2.1 Responds:} Based on the provided historical data, this region in Staten Island seems to experience fluctuating yet relatively low bike traffic flow during the daytime on Tuesdays. The bike inflows range from 6 to 15 per 30-minute interval, while outflows range from 4 to 15.Given the region covers education, cultural, commercial, and transportation facilities within a 1km radius, the bike traffic likely corresponds to people traveling to and from these points of interest. The fluctuations could reflect class/work schedules and transportation needs throughout the day.For the evening prediction period of 6PM to 11:30PM, we may expect some further fluctuations but generally declining bike traffic as the day turns to night. People would be finishing up activities and heading home. I would predict the following bike inflows and outflows for the next 12 time steps: \textbf{[10 8 7 6 5 4 3 2 1 1 1 1]} and \textbf{[8 6 5 4 3 3 2 2 1 1 1 1]}... \\
    \midrule
    \textbf{UrbanGPT Responds:} Based on the given information, the predictions (\textcolor{tblue}{predicted tokens}) of bike inflow and outflow in this region are \textcolor{tblue}{<ST\_PRE>} \textbf{[20 21 22 21 17 12 9 7 6 5 5 5]} and \textcolor{tblue}{<ST\_PRE>} \textbf{[11 10 8 7 6 6 5 4 4 3 3 2]}.
    \\
    \bottomrule
  \end{tabular}
  \label{tab:case1}
\end{table*}

\vspace{1cm}

\begin{table*}[htbp]
  \centering
  \caption{Zero-shot predictions of different LLMs for taxi flow in NYC with the provided instructions (without spatial information).}
  \vspace{-0.1cm}
  \begin{tabular}{p{17.3cm}}
    \toprule
    \textbf{Instructions:} Given the historical data for taxi flow over 12 time steps in a specific region of New York City, the recorded taxi inflows are [111 86 90 78 49 75 46 32 20 14 9 9], and the recorded taxi outflows are [100 91 58 39 47 47 19 22 21 9 16 10]. The recording time of the historical data is 'January 5, 2020, 20:30, Sunday to January 6, 2020, 02:00, Monday, with data points recorded at 30-minute intervals'. \textcolor{tgreen}{No description is available for this region.} Now we want to predict the taxi inflow and outflow for the next 12 time steps during the time period of 'January 6, 2020, 02:30, Monday to January 6, 2020, 08:00, Monday, with data points recorded at 30-minute intervals'... \\
    \midrule
    \textbf{Ground true:} Based on the given information, the predictions of taxi inflow and outflow in this region are \textbf{[9 3 3 4 6 3 30 22 58 74 91 110]} and \textbf{[5 10 7 15 11 19 43 46 90 146 213 187]}. \\
    \midrule
    \textbf{Llama-2-70b Responds:} Based on the provided historical data, I have analyzed the traffic patterns... Taxi Inflow: \textbf{[105 90 80 70 60 55 45 35 25 15 10 5]} Taxi Outflow: \textbf{[95 85 70 60 50 45 35 25 15 10 5 0]}. The predictions are based on the following factors... \\
    \midrule
    \textbf{ChatGPT Responds:} To generate predictions for the next 12 time steps, we can use the average values as a baseline...Predicted taxi inflows: \textbf{[53.33 53.33... 53.33]}. Predicted taxi outflows: \textbf{[40.83 40.83... 40.83]}... \\
    \midrule
    \textbf{Claude-2.1 Responds:} The data covers evening hours on a Sunday and early morning hours on a Monday. We can expect lower taxi demand during later Sunday evening, with some pickup around end-of-weekend times and early Monday commute times. The region is unspecified, so we cannot infer any location-specific usage patterns. Inflow ranges from 49-111 taxis over the 30 minute intervals. Outflow ranges from 19-100 taxis. This indicates the region likely serves both residential and commercial purposes.\\
    Prediction: Inflows: \textbf{[25 15 10 8 7 9 15 25 35 40 45 50]}. Outflows: \textbf{[40 35 20 12 8 10 20 30 45 50 55 60]}... \\
    \midrule
    \textbf{UrbanGPT Responds:} Based on the given information, the predictions (\textcolor{tblue}{predicted tokens}) of taxi inflow and outflow in this region are \textcolor{tblue}{<ST\_PRE>} \textbf{[4 4 2 3 4 13 25 39 56 80 101 117]} and \textcolor{tblue}{<ST\_PRE>} \textbf{[6 4 2 2 4 13 24 39 56 80 100 115]}.
    \\
    \bottomrule
  \end{tabular}
  \label{tab:case2}
\end{table*}

\end{document}